\documentclass[letterpaper]{article}
\usepackage{aaai25}
\nocopyright
\usepackage{times}
\usepackage{helvet}
\usepackage{courier}
\usepackage[hyphens]{url}
\usepackage{graphicx}
\usepackage{natbib}
\usepackage{caption}
\usepackage{algorithm}
\usepackage{newfloat}
\usepackage{listings}
\DeclareCaptionStyle{ruled}{labelfont=normalfont,labelsep=colon,strut=off}
\floatstyle{ruled}
\newfloat{listing}{tb}{lst}{}
\floatname{listing}{Listing}
\usepackage{booktabs}
\usepackage{algpseudocode}
\usepackage{subcaption}
\usepackage{amsmath}
\usepackage{amssymb}
\usepackage{tikz}

\title{Optimizing Interpretable Decision Tree Policies for Reinforcement Learning}
\author {
    Dani\"el Vos\textsuperscript{},
    Sicco Verwer\textsuperscript{}
}
\affiliations {
    \textsuperscript{}Delft University of Technology\\
    d.a.vos@tudelft.nl, s.e.verwer@tudelft.nl
}

\begin{document}

\maketitle

\begin{abstract}
Reinforcement learning techniques leveraging deep learning have made tremendous progress in recent years. However, the complexity of neural networks prevents practitioners from understanding their behavior. Decision trees have gained increased attention in supervised learning for their inherent interpretability, enabling modelers to understand the exact prediction process after learning. This paper considers the problem of optimizing interpretable decision tree policies to replace neural networks in reinforcement learning settings. Previous works have relaxed the tree structure, restricted to optimizing only tree leaves, or applied imitation learning techniques to approximately copy the behavior of a neural network policy with a decision tree. We propose the Decision Tree Policy Optimization (DTPO) algorithm that directly optimizes the complete decision tree using policy gradients. Our technique uses established decision tree heuristics for regression to perform policy optimization. We empirically show that DTPO is a competitive algorithm compared to imitation learning algorithms for optimizing decision tree policies in reinforcement learning.
\end{abstract}

\section{Introduction}

In recent years, many successful (deep) neural network-based techniques have been proposed for reinforcement learning~\cite{mnih2015human,schulman2017proximal}. However, due to the size and structure of these models, the resulting policies cannot be interpreted, which limits their use in real-life applications.
Decision trees are a popular model type in supervised learning as they can be directly interpreted~\cite{lipton2018mythos} and efficiently learned with heuristics~\cite{breiman1984classification,quinlan1986induction}.
Using decision trees as reinforcement learning policies is, therefore, a promising research direction. Unfortunately, decision trees are difficult to optimize for reinforcement learning because existing algorithms require models to be differentiable, which is not possible for decision trees due to their discontinuity.

Some methods have been proposed for optimizing decision tree policies by working around their non-differentiability, but they come with caveats. VIPER~\cite{bastani2018verifiable} first trains a Deep Q-Network~\cite{mnih2015human} and then extracts a decision tree from the neural network using imitation learning. While this often results in performant decision trees, it is time-consuming and relies on a good teacher model. Other methods use knowledge of the Markov Decision Process (MDP) underlying the environment, e.g., Iterative bounding MDPs~\cite{topin2021iterative} and Optimal MDP Decision Trees~\cite{vos2023optimal}, which means that they cannot solve most reinforcement learning problems. In another line of work, the non-differentiable parts of decision tree models have been relaxed~\cite{silva2020optimization,paleja2022learning} or restricted~\cite{likmeta2020combining} to enable gradient-based optimization. These relaxations or restrictions are later removed to obtain a crisp decision tree, which can perform much worse than its soft (non-crisp) counterpart. 
Existing methods that learn trees with a crisp structure ~\cite{gupta2015policy,roth2019conservative} do so using a greedy splitting procedure that can never update the tree structure again once creating a branch node, making it hard to recover from mistakes.

\begin{figure*}[tb]
    \centering
    \includegraphics[width=\textwidth]{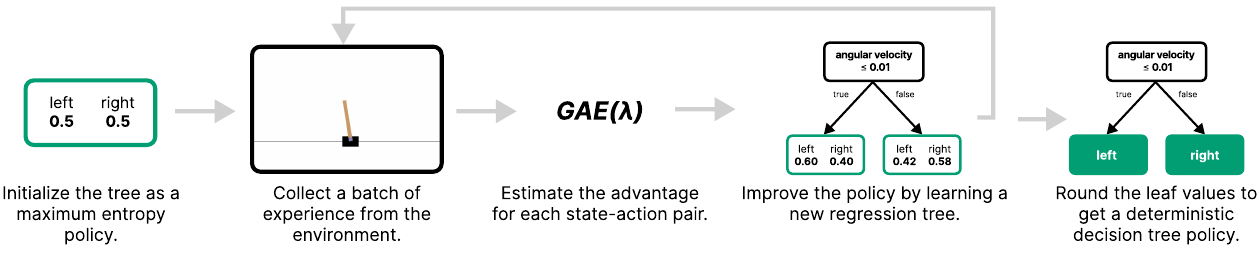}
    \caption{High-level overview of the DTPO algorithm. The tree is initialized as a single leaf with equal probability for each action and iteratively refined by optimizing the loss with regression tree heuristics on batches of environment experience. In the end, we round the leaf values to obtain an interpretable deterministic policy.}
    \label{fig:dtpo-overview}
\end{figure*}

We propose Decision Tree Policy Optimization (DTPO), an iterative method that uses regression tree learning heuristics that can incrementally improve decision trees for a differentiable loss function. 
DTPO is inspired by the success of PPO~\cite{schulman2017proximal}, a policy gradient style method that has gained widespread attention for its strong performance and relative simplicity.
A high-level overview of the method is given in Figure~\ref{fig:dtpo-overview}. DTPO allows us to use gradient-based algorithms to optimize reinforcement learning policies without imitation learning or altering the model class. We evaluate DTPO on several control tasks and discrete MDPs and compare the performance to the state-of-the-art method VIPER. DTPO performs competitively with VIPER, and in 4 of the 17 tested environments, one of the decision tree methods outperforms the neural network-based methods. To the best of our knowledge, DTPO is the first method capable of optimizing arbitrary differentiable loss functions for decision trees in reinforcement learning and is significantly simpler than previous works. 

\section{Background and Related Work}
We briefly introduce the topics of decision tree optimization, reinforcement learning, and techniques for optimizing decision tree policies in reinforcement learning.

\subsection{Decision Trees in Supervised Learning}
Decision trees are relatively simple models that explicitly model a set of rules with predictions in the form of a tree. An example of a small decision tree is given in Figure~\ref{fig:dtpo-overview} on the right side. Recently, such models have gained increased attention due to their human interpretability when limited in size~\cite{lipton2018mythos}. When learning decision trees, it is common to use greedy heuristics based on Classification and Regression Trees~\cite{breiman1984classification} and ID3~\cite{quinlan1986induction}; these algorithms recursively split up the dataset with branching nodes until ready to create a leaf node and make a prediction. In this work, we use the multi-output regression trees as implemented by Scikit-learn~\cite{pedregosa2011scikit}. These trees predict multiple continuous values within each leaf and heuristically minimize mean squared error using a greedy splitting algorithm.

\subsection{Reinforcement Learning}

Reinforcement learning is the problem of finding a policy that maximizes the expected sum of rewards in an unknown Markov Decision Process (MDP) by repeatedly acting and observing the outcomes. Algorithms that solve this problem without explicitly modeling the MDP are called model-free, and there are two main styles of learning algorithms: Q-learning and policy gradients. In Q-learning, the agent attempts to learn the Q-value function, which maps states and actions to a value that represents how good it is to take the action in the given state. In this work, we will compare to Deep Q-Networks~\cite{mnih2015human}, models that solve reinforcement learning problems by approximating the Q-value function with a neural network. 

Policy gradient style algorithms~\cite{sutton1999policy} directly optimize a policy that maps states to action probabilities by running the policy in the environment, estimating the advantage of performing its predicted actions, and updating the policy to do more or less of those actions depending on the advantage estimates. A major difference between Q-learning and policy gradient methods is that Q-learning usually deviates from its current policy to learn Q-values for the whole state space, while policy gradient methods only improve the policy in states that are reached by that policy. Recent work~\cite{vos2023optimal} has shown that due to the limited capacity of decision trees, it can be better to only optimize for the part of the state space reached by the policy, which motivates the choice of searching over policies versus Q-learning in this paper.

\subsubsection{Proximal Policy Optimization}
In this paper, we base our algorithm on Proximal Policy Optimization (PPO)~\cite{schulman2017proximal}, which is an established policy gradient-style method for reinforcement learning. 
PPO simultaneously trains a policy that determines the actions to take in the environment and a value function that predicts the value of the agent being in a state, both of which are modeled with a neural network. The value estimates are used to improve learning of the policy.
Specifically, PPO maximizes the following objective for the policy where $R_t(\theta) = \frac{\pi_\theta(a_t | s_t)}{\pi_{\theta_\text{old}}(a_t | s_t)}$ is the ratio of probabilities $\pi_\theta(a_t | s_t)$ assigned to the taken action $a_t$ in state $s_t$ between the updated and old policy:
{
\small
\begin{equation} \label{eq:ppo-objective}
    L^\text{CLIP}(\theta) = \mathbb{E}_t\left[ \min(R_t(\theta) \hat{A}_t, \text{clip}(R_t(\theta), 1{-}\epsilon, 1{+}\epsilon) \hat{A}_t) \right].
\end{equation}
}
\noindent Here $\theta$ represents the policy network parameters, $\hat{A_t}$ the estimated advantage of environment interaction at time $t$ and $\epsilon$ the clipping hyperparameter, which is typically set to $\epsilon = 0.2$. By taking the minimum of the clipped and unclipped terms, the agent does not get rewarded for changes to the action probabilities beyond a ratio of $[1-\epsilon,1+\epsilon]$. To compute the advantage estimates $\hat{A_t}$, PPO uses a truncated version of Generalized Advantage Estimation~\cite{schulman2015high,mnih2016asynchronous}:
\begin{align} \label{eq:gae}
    &\hat{A} = \delta_t + (\gamma \lambda) \delta_{t + 1} + ... + (\gamma \lambda)^{T - t + 1} \delta_{T - 1}, \\
    &\text{where} \;\; \delta_t = r_t + \gamma V(s_{t+1}) - V(s_{t}). \nonumber
\end{align}
Here $T$ is the total number of timesteps, $r_t$ the reward at time $t$, $\gamma$ the discount factor, $\lambda$ the mixing factor, and $V$ the value function neural network. Both $\gamma$ and $\lambda$ weigh the importance of the current timestep on future reward and are typically set to a number close but less than 1 (typically $\gamma = 0.99$ and $\lambda = 0.95$). The advantage estimates are typically normalized to zero mean and unit variance.

The value neural network is trained by minimizing the squared error loss between observed and predicted returns and is also often clipped to prevent large updates:
\begin{multline} \label{eq:clipped-value-loss}
    L^\text{VALUE}(\theta) = \max \bigl((V_{\theta}(s_t) - V_t)^2,\\
    V_{\theta_\text{old}(s_t)} + \text{clip}(V_{\theta}(s_t) - V_{\theta_\text{old}}(s_t), -\epsilon, \epsilon) - V_t)^2 \bigl).
\end{multline}
To optimize decision trees with PPO, we will replace the policy neural network with a decision tree but still use a neural network for the value function, as this approximator is only used to improve the optimization process and does not affect the interpretability of the policy model.

\subsection{Optimizing Decision Tree Policies}
We briefly introduce the existing works on optimizing decision tree policies within reinforcement learning.
Other methods for interpretable or post-hoc explainable methods for reinforcement learning are discussed in~\cite{glanois2021survey} and \cite{milani2022survey}.

Some methods have been proposed before that relax or restrict the decision tree learning problem to allow gradient updates. Policy tree~\cite{gupta2015policy} is a method that uses linear models as tree leaves together with a greedy splitting procedure to train decision tree policies with policy gradient techniques. While this algorithm trains well, the resulting tree model consists of multiple different linear models in the leaves, which makes it difficult to interpret the predictive behavior and can never backtrack once a decision node has been created. Similarly, Conservative Q-Improvement~\cite{roth2019conservative} greedily expands a decision tree to improve the Q-values, but the algorithm cannot update the tree structure. Differentiable decision trees~\cite{silva2020optimization} relax the property that each branch node selects one feature, that each leaf predicts a single action, and that only one branch is taken at a time; the resulting 'soft' decision tree is then optimizable with PPO. Although the soft decision tree can be optimized, the model usually incurs a significant drop in performance when discretizing into a 'crisp' decision tree. Paleja et al.~[\citeyear{paleja2022learning}] improve this discretization error by using a different relaxation technique, but to achieve small, performant trees, the algorithm uses linear functions in the leaves, which makes the models harder to interpret. Likmeta et al.~[\citeyear{likmeta2020combining}] propose fixing the branching nodes of the decision tree by using expert knowledge and optimizing the differentiable leaf values using gradient descent.

Iterative Bounding MDPs~\cite{topin2021iterative} is a method that extends known \emph{factored} MDPs~\cite{boutilier1995exploiting} with tree-building actions such that neural networks can be used to optimize them. Optimal MDP Decision Trees~\cite{vos2023optimal} is a method for identifying the globally optimal decision tree policy for a known discrete MDP. Since both these methods require a specification of the complete MDP, they cannot be applied directly to the reinforcement learning setting. VIPER~\cite{bastani2018verifiable} uses an improved imitation learning algorithm to extract a decision tree from a more complex teacher model trained with Q-learning, such as a Deep Q-Network (DQN)~\cite{mnih2015human}. While VIPER works well in most environments, it requires a good Q-learning teacher, and in some environments, Q-learning is unfavorable. dtControl~\cite{ashok2020dtcontrol} exactly converts controllers into decision trees for verification, but the resulting trees are usually too large to be interpreted. To the best of our knowledge, no published policy gradient-style algorithm can directly optimize whole decision trees for reinforcement learning.

\begin{algorithm*}[tb]
    \caption{DTPO: Decision Tree Policy Optimization}\label{alg:DTPO}
    \begin{algorithmic}[1]
        \State Initialize $\pi$ as a single leaf that takes any of the $n$ actions with equal probability: $\pi(s) = \left(\frac{1}{n}, \frac{1}{n}, ..., \frac{1}{n}\right) \; \forall s$
        \For{$i = 1,2, ... N$}
            \State Run policy $\pi$ in the environment for $t = 1...T$ timesteps collecting observations $s_t$, actions $a_t$, and rewards $r_t$
            \State Compute the $GAE(\lambda)$ advantage estimates $\hat{A}_t$ using Equation~\ref{eq:gae}, normalize them to zero mean and unit variance
            \State Define the log action probabilities at time $t$ as $l_t = \log(\pi_{i - 1}(s_t) + 10^{-8})$
            \State Fit regression tree $\pi'$ on observations $s_t$ with targets $Y_t = \sigma\left(l_t + \eta \nabla L^\text{DT}(l_t)\right)$ \Comment{See Equation~\ref{eq:loss-dt} for $L^\text{DT}$}
            \State Replace $\pi$ by $\pi'$ if it improves the $L^\text{DT}$ loss
            \State Once every 10 iterations: evaluate a deterministic copy of $\pi$
        \State Optimize the value function parameters with $L^\text{VALUE}$ $\theta$ loss, for $E$ epochs with batches of size $B$ using Adam
    \EndFor
    \State Return the best deterministic policy encountered during optimization.
    \end{algorithmic}
\end{algorithm*}

\section{DTPO: Decision Tree Policy Optimization}

In this section, we describe Decision Tree Policy Optimization (DTPO), an algorithm for optimizing decision tree policies in reinforcement learning settings. DTPO is based on PPO~\cite{schulman2017proximal}, a popular policy gradient-based technique for optimizing reinforcement learning policies represented by neural networks. Previous works have usually not applied gradient-based algorithms to decision trees as decision trees are not differentiable. We propose a method that makes incremental improvements to a decision tree policy without requiring the model to be differentiable, allowing us to use the policy gradient information during optimization. We first explain our technique for incremental updates using regression tree learning heuristics, followed by an explanation of how this technique is applied inside the DTPO algorithm.

\subsection{Incremental Regression Tree Improvement} \label{sec:iterative-tree-improvement}

A major challenge in decision tree optimization is that each decision node splits on a single feature and uses a hard threshold that sends samples to the left or right subtree. This property makes the prediction space discontinuous and also has the effect that changing a single decision node can lead to a very large change in predictions. Due to these issues, existing algorithms for incremental decision tree-based learning usually resort to ensembles, which comes at the cost of interpretability. For example, gradient boosting~\cite{mason1999boosting,hastie2009boosting} gradually reduces the loss by iteratively adding decision trees to its ensemble. These trees compensate for the loss of the preceding trees in the ensemble by predicting pseudo-residuals based on the gradient of the loss. We take inspiration from gradient boosting and propose a simple yet surprisingly effective method that optimizes a single decision tree for a differentiable loss function.

The main idea is to leverage existing regression tree learning heuristics and repurpose them to iteratively optimize regression trees for a differentiable loss function $L$ on a sufficiently large batch of samples $X$. By updating the targets of the learner according to the gradient of the loss in each iteration, we can use gradient information without having to differentiate through the decision tree. Define the loss as a differentiable function $L: \mathcal{Y} \rightarrow \mathbb{R}$, mapping a set of model predictions to a value to be minimized. Our method works as follows:
\begin{enumerate}
    \item initialize a decision tree $\mathcal{T}_0: \mathcal{X} \rightarrow \mathcal{Y}$ that maps samples $x \in X$ to (arbitrary) predictions $Y_0$,
    \item learn a regression tree $\mathcal{T}_{i}$ aiming to predict $\mathcal{T}_{i}(X) \approx Y_{i - 1} - \eta \nabla L(Y_{i - 1})$, with learning rate $\eta$,
    \item (optionally) repeat step 2 for $N$ iterations and return the best performing tree $\mathcal{T}_{\text{argmin}_i(L(Y_i))}$.
\end{enumerate}
Like gradient boosting, this method iteratively reduces the pseudo-residuals, but instead of adding a tree to the ensemble, the new tree represents the whole ensemble. While this method does not guarantee an improvement of the loss value in every step, we find that it works well in practice. The pseudocode for (multi-output) regression tree learning~\cite{breiman1984classification} is given in the appendix.

\subsection{Policy Optimization}

To perform policy optimization with decision trees, we replace the policy neural network that is typically used with a regression tree that predicts the probability of each action. We still use a neural network to learn the value function. We then perform gradient updates using the method outlined in the previous section, and we do not perform the optional step 3 as we find that it is not necessary for good performance. The algorithm's pseudocode is given in Algorithm~\ref{alg:DTPO}. An important requirement for this method is that the batch size is large enough to capture sufficient information about the tree of the previous iteration so as not to `forget' behavior. This is in contrast to neural network optimization, where information from previous batches is effectively stored in the model parameters and approximately persisted by limiting the magnitude of parameter updates. Therefore, in our experiments, we collect batches of $T = 10,000$ steps of experience similar to methods such as TRPO~\cite{schulman2015trust}. This number can be decreased when training very small trees or increased to, e.g., $T = 50,000$ for more complex environments.

While our algorithm is inspired by PPO, we make some changes to improve learning using decision trees. 
We only perform one update per batch on the decision tree's CLIP loss (Equation \ref{eq:ppo-objective}) instead of performing multiple epochs on minibatches, as is typically done in PPO.
This means that effectively, the decision tree is optimizing only the first term of the $L^\text{CLIP}$ loss (Equation~\ref{eq:ppo-objective}). We define this decision tree loss in terms of the logits $l$ as:
\begin{equation} \label{eq:loss-dt}
    L^\text{DT}(l) = \mathbb{E}_t\left[ \frac{\sigma(l_{t})_{a_t}}{\pi(a_t | s_t)} \hat{A}_t \right]
\end{equation}
where $\sigma$ is the softmax function $\sigma(\mathbf{x})_i = e^{x_i} / \sum_{j} e^{x_{j}}$ and $\hat{A}_t$ the advantage estimate at time $t$.
This loss is similar to the one used in vanilla policy gradients. Performing multiple updates using the $L^\text{CLIP}$ loss is possible but not required for finding high-quality deterministic policies. In DTPO, we train regression trees that directly predict probabilities, and we use softmax functions to make sure that these probabilities sum to one after gradient updates. It is also possible to predict logits with the regression trees and only apply the softmax function $\sigma$ when computing probabilities, but the regression tree learning heuristics work better for equally scaled outputs (such as probabilities).

As mentioned in the previous section, incremental regression tree improvement does not guarantee an improvement; therefore, we check the loss values before and after the procedure, and if the loss does not improve, we revert the update. To choose the final decision tree policy, we determinize it every 10 iterations and evaluate it based on a separate batch of samples. At the end of the optimization process, we return the policy with the best deterministic performance. Determinization is performed by replacing each leaf's action probabilities with a distribution that places all probability mass on the most probable action.

In every iteration, we also update the parameters $\theta$ of the neural network that we use to approximate the value function (the critic). This update minimizes PPO's clipped value loss (Equation~\ref{eq:clipped-value-loss}) with the Adam~\cite{kingma2014adam} optimizer for $E$ epochs on batches of $B$ samples.

\begin{table*}[tb]
\centering
\begin{tabular}{l|rr|rr|rr}
\toprule
 & \multicolumn{2}{c|}{number of:} & \multicolumn{2}{c}{decision tree policies} & \multicolumn{2}{c}{neural network policies} \\
Environment & features & actions & \multicolumn{1}{c}{VIPER} & \multicolumn{1}{c|}{DTPO} & \multicolumn{1}{c}{DQN} & \multicolumn{1}{c}{PPO} \\ \midrule
\multicolumn{7}{c}{Control tasks} \\ \midrule
Acrobot-v1 & 6 & 3 & -87.24 \tiny $\pm$ \hphantom{00}5.00 & \textbf{-85.39} \tiny $\pm$ \hphantom{00}0.87 & \textbf{-65.76} \tiny $\pm$ \hphantom{00}1.62 & -79.17 \tiny $\pm$ \hphantom{00}1.47 \\
CartPole-v1 & 4 & 2 & 367.10 \tiny $\pm$ 121.11 & \textbf{493.75} \tiny $\pm$ \hphantom{00}6.24 & 305.57 \tiny $\pm$ 112.01 & \textbf{492.62} \tiny $\pm$ \hphantom{00}4.38 \\
CartPoleSwingup & 5 & 2 & \textbf{444.02} \tiny $\pm$ \hphantom{0}18.43 & 413.62 \tiny $\pm$ \hphantom{0}52.34 & \textbf{599.08} \tiny $\pm$ 103.99 & 517.10 \tiny $\pm$ 170.67 \\
Pendulum-v1 (discrete) & 3 & 2 & \textbf{-207.49} \tiny $\pm$ \hphantom{0}56.20 & -628.49 \tiny $\pm$ \hphantom{00}8.27 & -153.76 \tiny $\pm$ \hphantom{00}3.66 & \textbf{-149.29} \tiny $\pm$ \hphantom{00}1.13 \\ \midrule
\multicolumn{7}{c}{Discrete OMDT environments} \\ \midrule
Blackjack & 3 & 2 & \textbf{-22.96} \tiny $\pm$ \hphantom{00}0.14 & -25.29 \tiny $\pm$ \hphantom{00}1.68 & \textbf{-22.47} \tiny $\pm$ \hphantom{00}0.23 & -24.56 \tiny $\pm$ \hphantom{00}0.45 \\
Frozenlake4x4 & 2 & 4 & \textbf{0.74} \tiny $\pm$ \hphantom{00}0.01 & 0.73 \tiny $\pm$ \hphantom{00}0.00 & \textbf{0.74} \tiny $\pm$ \hphantom{00}0.01 & 0.54 \tiny $\pm$ \hphantom{00}0.17 \\
Frozenlake8x8 & 2 & 4 & \textbf{0.85} \tiny $\pm$ \hphantom{00}0.04 & 0.42 \tiny $\pm$ \hphantom{00}0.23 & \textbf{0.84} \tiny $\pm$ \hphantom{00}0.03 & 0.82 \tiny $\pm$ \hphantom{00}0.01 \\
Frozenlake12x12 & 2 & 4 & \textbf{0.62} \tiny $\pm$ \hphantom{00}0.07 & 0.15 \tiny $\pm$ \hphantom{00}0.10 & \textbf{0.64} \tiny $\pm$ \hphantom{00}0.08 & 0.01 \tiny $\pm$ \hphantom{00}0.01 \\
InventoryManagement & 1 & 100 & \textbf{9659.39} \tiny $\pm$ 382.11 & 8605.77 \tiny $\pm$ 409.58 & 9658.49 \tiny $\pm$ 383.53 & \textbf{10152.71} \tiny $\pm$ \hphantom{00}8.02 \\
Navigation3D & 3 & 6 & 0.00 \tiny $\pm$ \hphantom{00}0.00 & \textbf{49.30} \tiny $\pm$ \hphantom{00}1.36 & 0.00 \tiny $\pm$ \hphantom{00}0.00 & \textbf{53.20} \tiny $\pm$ \hphantom{00}0.53 \\
SystemAdministrator1 & 8 & 9 & \textbf{1709.39} \tiny $\pm$ \hphantom{00}5.34 & 1610.00 \tiny $\pm$ \hphantom{0}63.53 & \textbf{1707.63} \tiny $\pm$ \hphantom{00}2.90 & 1676.10 \tiny $\pm$ \hphantom{00}1.68 \\
SystemAdministrator2 & 8 & 9 & \textbf{1727.08} \tiny $\pm$ 129.65 & 1678.17 \tiny $\pm$ \hphantom{0}17.77 & 1726.52 \tiny $\pm$ 116.23 & \textbf{1784.00} \tiny $\pm$ \hphantom{0}17.04 \\
SystemAdministratorTree & 7 & 8 & 4842.50 \tiny $\pm$ \hphantom{0}87.39 & \textbf{5411.78} \tiny $\pm$ 111.68 & 4840.33 \tiny $\pm$ \hphantom{0}89.07 & \textbf{5505.93} \tiny $\pm$ \hphantom{0}14.05 \\
TictactoeVsRandom & 27 & 9 & 0.67 \tiny $\pm$ \hphantom{00}0.02 & \textbf{0.86} \tiny $\pm$ \hphantom{00}0.03 & \textbf{0.98} \tiny $\pm$ \hphantom{00}0.00 & 0.97 \tiny $\pm$ \hphantom{00}0.01 \\
TigerVsAntelope & 4 & 5 & 0.27 \tiny $\pm$ \hphantom{00}0.14 & \textbf{1.00} \tiny $\pm$ \hphantom{00}0.00 & 0.19 \tiny $\pm$ \hphantom{00}0.07 & \textbf{1.00} \tiny $\pm$ \hphantom{00}0.00 \\
TrafficIntersection & 4 & 2 & \textbf{28.24} \tiny $\pm$ \hphantom{00}0.81 & 26.25 \tiny $\pm$ \hphantom{00}0.67 & \textbf{29.83} \tiny $\pm$ \hphantom{00}0.50 & -37.21 \tiny $\pm$ \hphantom{00}0.94 \\
Xor & 2 & 2 & 996.65 \tiny $\pm$ \hphantom{00}3.35 & \textbf{1000.00} \tiny $\pm$ \hphantom{00}0.00 & \textbf{996.69} \tiny $\pm$ \hphantom{00}3.31 & 487.92 \tiny $\pm$ \hphantom{0}29.89 \\
\bottomrule
\end{tabular}
\caption{Mean and standard errors of undiscounted returns averaged over 3 random seeds and evaluated with 1000 rollouts. The best method varies per environment. The relative performance of VIPER and DTPO is correlated with the relative performance of DQN and PPO. DTPO outperforms it on environments that favor policy gradient methods (e.g., \textit{Navigation3D}), while VIPER wins on environments that favor Q-learning (e.g., \textit{Frozenlake}). In the environments \textit{Frozenlake8x8} and \textit{CartPole-v1}, the decision tree policies outperform neural network policies.}
\label{tab:performance-comparison}
\end{table*}

\section{Results}

We compare DTPO's ability to optimize interpretable tree policies with VIPER~\cite{bastani2018verifiable}, a method based on imitating a Q-learning teacher such as a Deep Q-Network~\cite{mnih2015human}. To maintain interpretability, we limit the trees to 16 leaf nodes, and to quantify the cost of interpretability, we also compare against neural network-based PPO~\cite{schulman2017proximal} and DQN. We implemented DTPO in JAX~\cite{jax2018github} and the reinforcement learning environments in Gymnax~\cite{gymnax2022github}. The experiments ran on a Linux machine with 16 Intel Xeon CPU cores and 72 GB of RAM, each method running on a single core. See our code\footnote{\url{https://github.com/tudelft-cda-lab/DTPO}} and appendix for more details.

For DTPO we use the hyperparameters: $\eta = 1.0, \gamma = 0.99, \lambda = 0.95, T = 10,000$, and $N = 1,500$. We update the value function using the Adam optimizer for $E = 4$ epochs on batches of $B = 64$ samples. 
When learning neural network PPO policies and all value functions, we use a standard neural network with 2 layers of 64 neurons, tanh activation functions, and a learning rate of 2.5e-4.
The remaining hyperparameters are given in the appendix.

\subsection{Performance Comparison}

We compare the algorithms on a variety of environments, including a set of gymnasium classic control tasks, the MDPs implemented by OMDT~\cite{vos2023optimal}, and cartpole swingup~\cite{ha2017evolving}: a much more challenging cartpole variant. These environments were chosen because they have informative features that allow for interpretation, unlike image-based environments, which have features that are individual pixels. All environments use discrete action spaces. Since our primary focus is on policy quality and not sample efficiency, we train the neural network methods DQN and PPO for 10 million total timesteps. VIPER uses the DQN to extract a decision tree using additional environment rollouts, which, when assuming rollouts of 1000 steps, results in VIPER using approximately 4.7 million additional environment samples. DTPO uses 1,500 batches of 10,000 samples and, therefore, uses a total of approximately 15 million samples. In Table~\ref{tab:performance-comparison}, we list the mean undiscounted returns (the environments' objectives) that the different methods achieved.

When the DQN finds a good policy for an environment, VIPER often outperforms DTPO. However, Since VIPER relies on the DQN to optimize its policy, the method fails in environments such as \textit{Navigation3D} and \textit{TigerVsAntelope}, where the DQN finds a bad policy. Q-learning and policy gradient techniques often excel in different types of environments, as can be seen in the results of the neural network policies. Therefore, DTPO is useful when Q-learning techniques fail to achieve good performance. For instance, the \textit{TigerVsAntelope} game and \textit{TictactoeVsRandom} DTPO find significantly better policies. In some environments, a decision tree-based policy outperforms the neural network-based policies; for example, on \textit{Frozenlake8x8}, VIPER outperforms the neural networks by a small margin, and on \textit{CartPole-v1}, DTPO outperforms the neural network policies.

\begin{figure*}[tb]
    \centering
    \centering
    \includegraphics[width=0.95\textwidth]{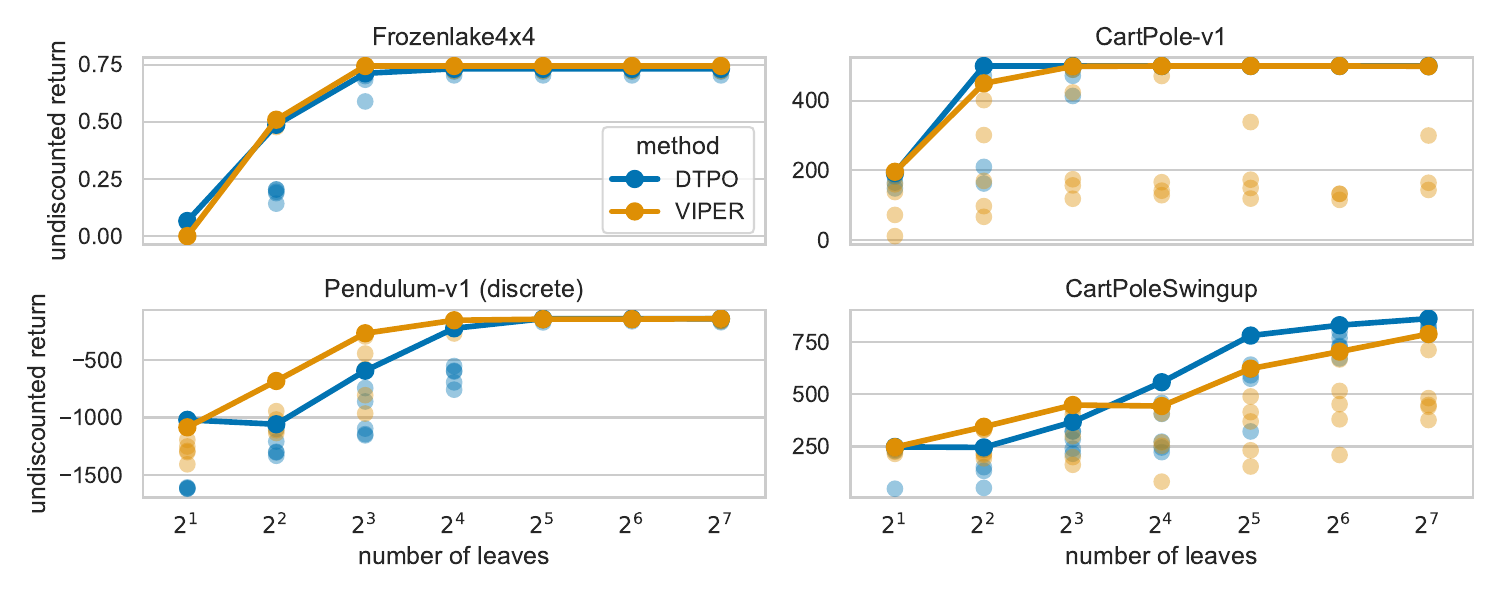}
    \caption{
    Undiscounted returns of policies with varying decision tree sizes. Each tree size was run with 6 random seeds, and best performing policies are highlighted. DTPO and VIPER perform similarly on average, but their performance varies depending on the environment. Simple environments such as \textit{Frozenlake4x4} reach optimal performance with policies of at most 8 leaves, while more complex problems like \textit{CartPoleSwingup} require at least 128 leaves to reach approximately optimal performance.}
    \label{fig:performance-vs-interpretability}
\end{figure*}

\subsection{Performance versus Interpretability}

An important consideration in interpretable machine learning is the tradeoff between simplicity and performance. To get a better understanding of the model complexity required to solve the environments under consideration, we experimented with decision tree policies with varying numbers of leaves. In Figure~\ref{fig:performance-vs-interpretability}, we visualize the best returns out of 6 random seeds on various environments and decision tree sizes. We trained a new DQN model for each random seed for VIPER and set $T = 50,000$ and $N = 500$ for DTPO.

One interesting question is how much a policy can be simplified before it loses performance. We find that we can get near-optimal performance on \textit{CartPole-v1} with 4 leaves, 8 leaves for \textit{FrozenLake4x4}, 16 leaves for \textit{Pendulum-v1}, and 64 leaves for \textit{CartPoleSwingup}. All of these are much smaller models than typical PPO neural networks with thousands of parameters, meaning we can find much simpler policies in these environments without losing performance.

We can also compare how well VIPER and DTPO optimize size-limited decision trees. When allowing trees of 128 leaves, both VIPER and DTPO perform roughly similarly, but there is a difference in performance for smaller trees. Specifically, VIPER performs better for trees of 8 and 16 leaves on \textit{Pendulum-v1}, while DTPO performs better for \textit{CartPoleSwingup} trees of 16, 32, 64, and 128 leaves. In general, DTPO and VIPER perform similarly on average, but relative performance varies depending on the environment.

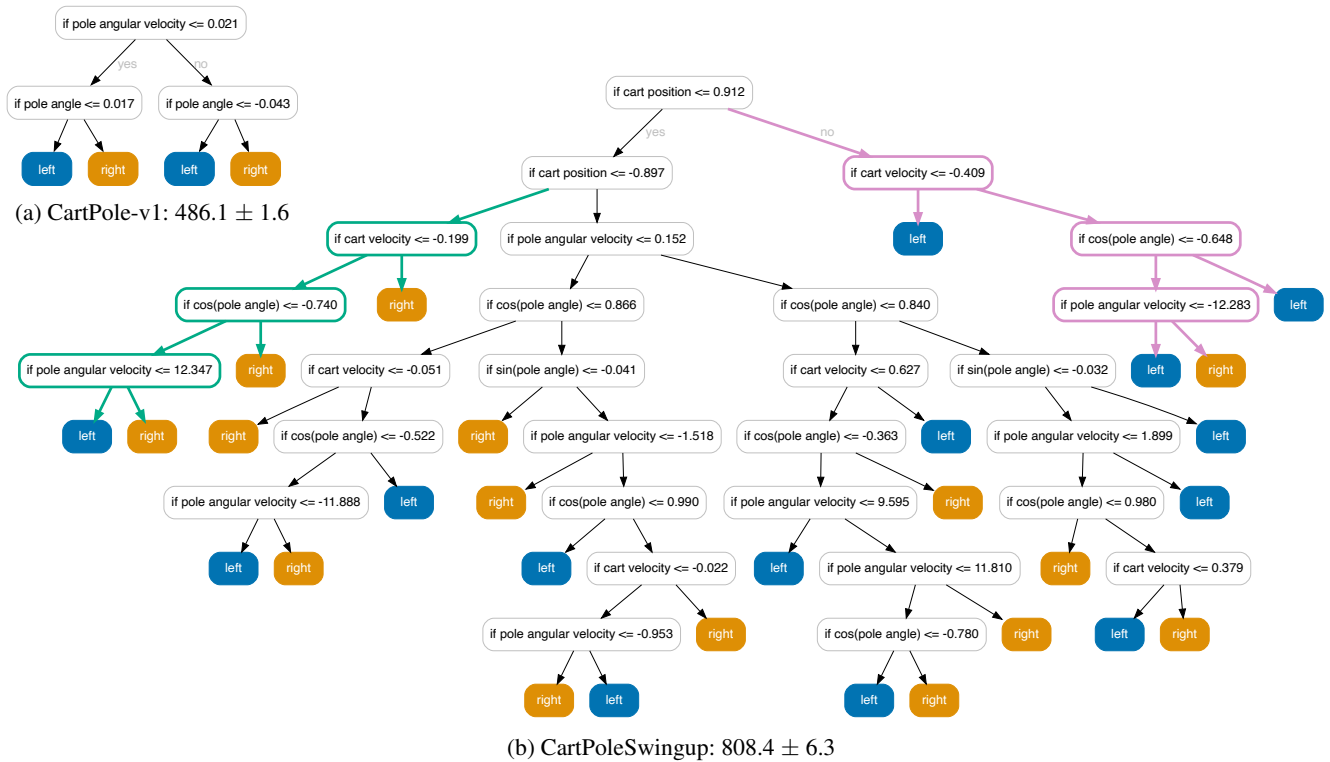
\begin{figure*}[tb]
    \centering
    \hfill
    \sbox0{
    \begin{subfigure}[t]{0.22\textwidth}
        \centering
        \includegraphics[width=0.99\textwidth]{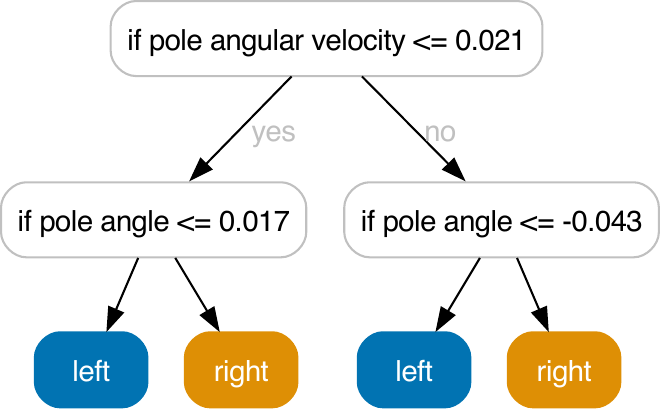}
        \caption{CartPole-v1: 486.1 $\pm$ 1.6}
        \label{fig:cartpole-policy}
    \end{subfigure}}
    \hfill
    \sbox1{
    \begin{subfigure}[t]{0.99\textwidth}
        \centering
        \includegraphics[width=0.99\textwidth]{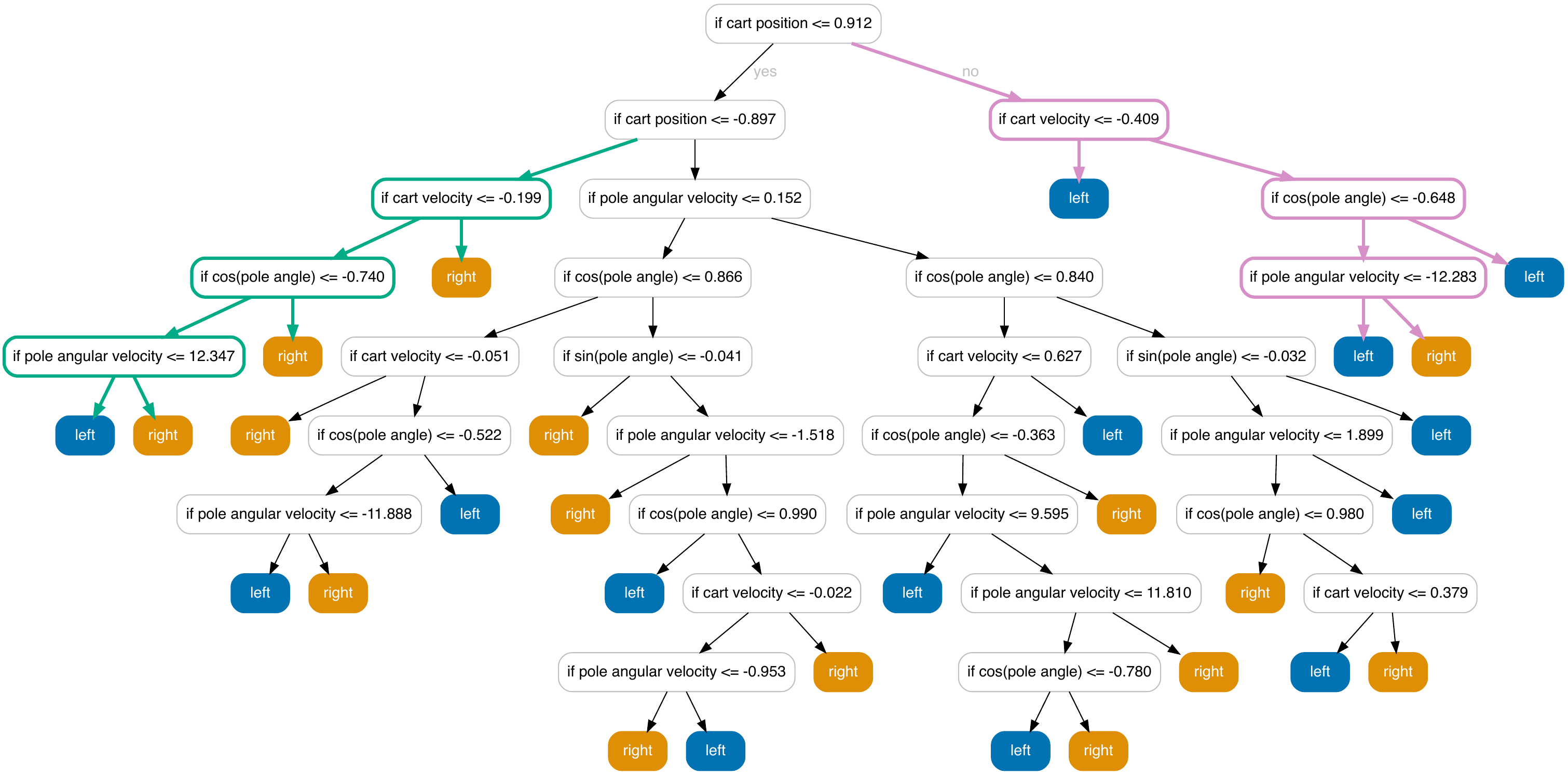}
        \caption{CartPoleSwingup: 808.4 $\pm$ 6.3}
        \label{fig:cartpoleswingup-policy}
    \end{subfigure}}
    \begin{tikzpicture}
        \node[inner sep=0pt] at (0,0) {\usebox1};
        \node[inner sep=0pt] at (-6.9,4) {\usebox0};
    \end{tikzpicture}
    \hfill
    \caption{
    DTPO policies for the simple \textit{CartPole-v1} and much harder \textit{CartPoleSwingup} environments, as environments increase in complexity, we need larger trees to achieve satisfactory performance. The explicit policy representation allows us to inspect them. This allows us to see, e.g., the symmetry in the \textit{CartPoleSwingup} policy (notice the highlighted subtrees). While the tree for \textit{CartPoleSwingup} is possibly too large to interpret exactly, one can still understand its parts and verify its properties. In contrast, a small PPO neural network policy typically consists of thousands of parameters and cannot easily be visualized.
    }
    \label{fig:learned-policies}
\end{figure*}

\subsection{Understanding Learned Policies}

A key feature of decision tree policies is that they can be directly interpreted. Therefore, we visualize DTPO policies for the CartPole-v1 and CartPoleSwingup environments in Figure~\ref{fig:learned-policies}. To determine the policy's predicted action for a state's observation, one starts at the top node of the tree and compares the observation values to the node's predicate, continuing to the left/right subtree if the predicate is true/false. The leaves determine the action to be taken by the policy. We analyze each of the visualized decision tree policies below.

\subsubsection{CartPole-v1}

In \textit{CartPole-v1}, the agent starts with the pole upright and is rewarded with 1 point per iteration, under the condition that the pole is kept upright and the cart does not exit the screen. A policy that successfully balances the pole is visualized in Figure~\ref{fig:cartpole-policy} and can be easily interpreted in terms of the 4 paths through the tree:
\begin{itemize}
    \setlength\itemsep{0mm}
    \item If the pole is turning to the left and angled to the left, move \textbf{left}. This will accelerate the pole to the right, balancing it.
    \item If the pole is turning to the right and angled to the right, move \textbf{right}. This will accelerate the pole to the left, balancing it.
    \item Since the left and right branches of the root node are almost equal, they can be interpreted together: If the pole is not on the same side as it is turning, \textbf{keep moving in the direction that the pole is on}. This will keep accelerating the pole in the direction it is moving.
\end{itemize}

\subsubsection{CartPoleSwingup}

\definecolor{newGreen}{rgb}{0, 0.67, 0.525}
\definecolor{newPink}{rgb}{0.84, 0.56, 0.78}

The \textit{CartPoleSwingup} environment is similar to CartPole-v1, but the pole does not start upright, significantly complicating the problem. Although the policy in Figure~\ref{fig:cartpoleswingup-policy} might be too large to completely interpret, we can still analyze its parts. Note that although this tree was trained with a size limit of 32 leaf nodes, the resulting tree only has 29 nodes. This is because we post-process every tree (also for VIPER) by replacing nodes whose children are all leaves that predict the same action with a single leaf. This improves interpretability without affecting the policy.

In the CartPoleSwingup problem, the agent is rewarded not only based on how vertical the pole is but also on how close the cart is to the center of the screen. Therefore, we can see the first nodes split on the \textit{cart position}, dividing the space into three regions:
\begin{itemize}
    \item \textit{cart position} $\leq -0.897$ (the cart is on the left, {\color{newGreen} green}): Move the cart back to the \textbf{right} unless the cart is moving left, the pole is angled far to the left, and angle is moving to the left, then move \textbf{left} to prevent the pole from falling.
    \item \textit{cart position} $> 0.912$ (the cart is on the right, {\color{newPink} pink}): Move the cart back to the \textbf{left} unless the cart is moving right, the pole is angled far to the right, and angle is moving to the right, then move \textbf{right} to prevent the pole from falling.
    \item \textit{cart position} $\in (-0.897, 0.912]$ (the cart is in the center): Perform \textbf{left} and \textbf{right} actions according to a policy based on pole angle, angular velocity, and cart velocity to try to get/keep the pole upright. For brevity, we do not analyze this subtree further.
\end{itemize}
Even though we do not force the tree to learn symmetrical behavior, we can see that there is a logical symmetry in the green and pink subtrees of the policy.

\section{Discussion}

We presented DTPO, a method that can directly optimize decision tree policies for reinforcement learning. By leveraging a method based on regression tree learning heuristics to update trees with gradient information, we were able to apply policy gradient optimization to decision trees. The resulting policies can be made small enough to be human-interpreted and performed competitively compared to existing algorithms that extract decision trees from neural network policies. Our experiments on classic control tasks and discrete MDPs demonstrate that small decision trees can sometimes perform as well as neural networks.

Although DTPO performed well in our benchmark, there are still limitations that need to be addressed. While neural networks can efficiently incrementally learn from small batches of data, our decision trees require a large batch size to function. This is because we re-learn a tree in each iteration, which means that, to avoid forgetting previous experience, the batch of experience must be large enough to hold the information of the previous tree. During development, we experimented with learning rates between 0.01 and 10 and batch sizes between 5,000 and 100,000. We also experimented with updating the policy multiple times per iteration. We noticed that our method worked well for various learning rates and for batches of 10,000 samples. In more complex environments, performance could be improved by using batches of 50,000 samples. Although the batch size influences sample efficiency, DTPO still has good runtime properties. This is because batches of 100,000 samples can be collected within seconds using the Gymnax environments. 

A common limitation in function approximation-based algorithms for reinforcement learning is that they find local optima, and this also applies to DTPO. Mechanisms that memorize and replay previous experience could help improve DTPO's sample efficiency, and additional exploration techniques could prevent the problem of local optima that policy gradient methods often suffer from.

DTPO provides a promising way to use gradient information inside of non-differentiable learners, such as decision trees. Future work might apply this idea to different loss functions in reinforcement or supervised learning that were previously hard to optimize. We also aim to leverage the fact that policy gradient techniques such as DTPO can be adapted for environments with continuous action spaces with relative ease. Lastly, recent works have proposed efficient methods for decision tree controller verification. We plan to use DTPO to train decision tree policies for new tasks and verify their safety properties.

\bibliography{aaai25}

\begin{thebibliography}{28}
\providecommand{\natexlab}[1]{#1}

\bibitem[{Ashok et~al.(2020)Ashok, Jackermeier, Jagtap, K{\v{r}}et{\'\i}nsk{\`y}, Weininger, and Zamani}]{ashok2020dtcontrol}
Ashok, P.; Jackermeier, M.; Jagtap, P.; K{\v{r}}et{\'\i}nsk{\`y}, J.; Weininger, M.; and Zamani, M. 2020.
\newblock dtControl: Decision tree learning algorithms for controller representation.
\newblock In \emph{Proceedings of the 23rd international conference on hybrid systems: Computation and control}, 1--7.

\bibitem[{Bastani, Pu, and Solar-Lezama(2018)}]{bastani2018verifiable}
Bastani, O.; Pu, Y.; and Solar-Lezama, A. 2018.
\newblock Verifiable reinforcement learning via policy extraction.
\newblock \emph{Advances in neural information processing systems}, 31.

\bibitem[{Boutilier et~al.(1995)Boutilier, Dearden, Goldszmidt et~al.}]{boutilier1995exploiting}
Boutilier, C.; Dearden, R.; Goldszmidt, M.; et~al. 1995.
\newblock Exploiting structure in policy construction.
\newblock In \emph{IJCAI}, volume~14, 1104--1113.

\bibitem[{Bradbury et~al.(2018)Bradbury, Frostig, Hawkins, Johnson, Leary, Maclaurin, Necula, Paszke, Vander{P}las, Wanderman-{M}ilne, and Zhang}]{jax2018github}
Bradbury, J.; Frostig, R.; Hawkins, P.; Johnson, M.~J.; Leary, C.; Maclaurin, D.; Necula, G.; Paszke, A.; Vander{P}las, J.; Wanderman-{M}ilne, S.; and Zhang, Q. 2018.
\newblock {JAX}: composable transformations of {P}ython+{N}um{P}y programs.

\bibitem[{Breiman et~al.(1984)Breiman, Friedman, Stone, and Olshen}]{breiman1984classification}
Breiman, L.; Friedman, J.; Stone, C.; and Olshen, R. 1984.
\newblock \emph{Classification and Regression Trees}.
\newblock Taylor \& Francis.
\newblock ISBN 9780412048418.

\bibitem[{Glanois et~al.(2021)Glanois, Weng, Zimmer, Li, Yang, Hao, and Liu}]{glanois2021survey}
Glanois, C.; Weng, P.; Zimmer, M.; Li, D.; Yang, T.; Hao, J.; and Liu, W. 2021.
\newblock A survey on interpretable reinforcement learning.
\newblock \emph{arXiv preprint arXiv:2112.13112}.

\bibitem[{Gupta, Talvitie, and Bowling(2015)}]{gupta2015policy}
Gupta, U.~D.; Talvitie, E.; and Bowling, M. 2015.
\newblock Policy tree: Adaptive representation for policy gradient.
\newblock In \emph{Proceedings of the AAAI Conference on Artificial Intelligence}, volume~29.

\bibitem[{Ha(2017)}]{ha2017evolving}
Ha, D. 2017.
\newblock Evolving Stable Strategies.
\newblock \emph{blog.otoro.net}.

\bibitem[{Hastie et~al.(2009)Hastie, Tibshirani, Friedman, Hastie, Tibshirani, and Friedman}]{hastie2009boosting}
Hastie, T.; Tibshirani, R.; Friedman, J.; Hastie, T.; Tibshirani, R.; and Friedman, J. 2009.
\newblock Boosting and additive trees.
\newblock \emph{The elements of statistical learning: data mining, inference, and prediction}, 337--387.

\bibitem[{Kingma and Ba(2014)}]{kingma2014adam}
Kingma, D.~P.; and Ba, J. 2014.
\newblock Adam: A method for stochastic optimization.
\newblock \emph{arXiv preprint arXiv:1412.6980}.

\bibitem[{Lange(2022)}]{gymnax2022github}
Lange, R.~T. 2022.
\newblock {gymnax}: A {JAX}-based Reinforcement Learning Environment Library.

\bibitem[{Likmeta et~al.(2020)Likmeta, Metelli, Tirinzoni, Giol, Restelli, and Romano}]{likmeta2020combining}
Likmeta, A.; Metelli, A.~M.; Tirinzoni, A.; Giol, R.; Restelli, M.; and Romano, D. 2020.
\newblock Combining reinforcement learning with rule-based controllers for transparent and general decision-making in autonomous driving.
\newblock \emph{Robotics and Autonomous Systems}, 131: 103568.

\bibitem[{Lipton(2018)}]{lipton2018mythos}
Lipton, Z.~C. 2018.
\newblock The mythos of model interpretability: In machine learning, the concept of interpretability is both important and slippery.
\newblock \emph{Queue}, 16(3): 31--57.

\bibitem[{Mason et~al.(1999)Mason, Baxter, Bartlett, and Frean}]{mason1999boosting}
Mason, L.; Baxter, J.; Bartlett, P.; and Frean, M. 1999.
\newblock Boosting algorithms as gradient descent.
\newblock \emph{Advances in neural information processing systems}, 12.

\bibitem[{Milani et~al.(2022)Milani, Topin, Veloso, and Fang}]{milani2022survey}
Milani, S.; Topin, N.; Veloso, M.; and Fang, F. 2022.
\newblock A survey of explainable reinforcement learning.
\newblock \emph{arXiv preprint arXiv:2202.08434}.

\bibitem[{Mnih et~al.(2016)Mnih, Badia, Mirza, Graves, Lillicrap, Harley, Silver, and Kavukcuoglu}]{mnih2016asynchronous}
Mnih, V.; Badia, A.~P.; Mirza, M.; Graves, A.; Lillicrap, T.; Harley, T.; Silver, D.; and Kavukcuoglu, K. 2016.
\newblock Asynchronous methods for deep reinforcement learning.
\newblock In \emph{International conference on machine learning}, 1928--1937. PMLR.

\bibitem[{Mnih et~al.(2015)Mnih, Kavukcuoglu, Silver, Rusu, Veness, Bellemare, Graves, Riedmiller, Fidjeland, Ostrovski et~al.}]{mnih2015human}
Mnih, V.; Kavukcuoglu, K.; Silver, D.; Rusu, A.~A.; Veness, J.; Bellemare, M.~G.; Graves, A.; Riedmiller, M.; Fidjeland, A.~K.; Ostrovski, G.; et~al. 2015.
\newblock Human-level control through deep reinforcement learning.
\newblock \emph{nature}, 518(7540): 529--533.

\bibitem[{Paleja et~al.(2022)Paleja, Niu, Silva, Ritchie, Choi, and Gombolay}]{paleja2022learning}
Paleja, R.~R.; Niu, Y.; Silva, A.; Ritchie, C.; Choi, S.; and Gombolay, M.~C. 2022.
\newblock Learning Interpretable, High-Performing Policies for Autonomous Driving.
\newblock \emph{Robotics: Science and Systems XVIII}.

\bibitem[{Pedregosa et~al.(2011)Pedregosa, Varoquaux, Gramfort, Michel, Thirion, Grisel, Blondel, Prettenhofer, Weiss, Dubourg et~al.}]{pedregosa2011scikit}
Pedregosa, F.; Varoquaux, G.; Gramfort, A.; Michel, V.; Thirion, B.; Grisel, O.; Blondel, M.; Prettenhofer, P.; Weiss, R.; Dubourg, V.; et~al. 2011.
\newblock Scikit-learn: Machine learning in Python.
\newblock \emph{the Journal of machine Learning research}, 12: 2825--2830.

\bibitem[{Quinlan(1986)}]{quinlan1986induction}
Quinlan, J.~R. 1986.
\newblock Induction of decision trees.
\newblock \emph{Machine learning}, 1: 81--106.

\bibitem[{Roth et~al.(2019)Roth, Topin, Jamshidi, and Veloso}]{roth2019conservative}
Roth, A.~M.; Topin, N.; Jamshidi, P.; and Veloso, M. 2019.
\newblock Conservative q-improvement: Reinforcement learning for an interpretable decision-tree policy.
\newblock \emph{arXiv preprint arXiv:1907.01180}.

\bibitem[{Schulman et~al.(2015{\natexlab{a}})Schulman, Levine, Abbeel, Jordan, and Moritz}]{schulman2015trust}
Schulman, J.; Levine, S.; Abbeel, P.; Jordan, M.; and Moritz, P. 2015{\natexlab{a}}.
\newblock Trust region policy optimization.
\newblock In \emph{International conference on machine learning}, 1889--1897. PMLR.

\bibitem[{Schulman et~al.(2015{\natexlab{b}})Schulman, Moritz, Levine, Jordan, and Abbeel}]{schulman2015high}
Schulman, J.; Moritz, P.; Levine, S.; Jordan, M.; and Abbeel, P. 2015{\natexlab{b}}.
\newblock High-dimensional continuous control using generalized advantage estimation.
\newblock \emph{arXiv preprint arXiv:1506.02438}.

\bibitem[{Schulman et~al.(2017)Schulman, Wolski, Dhariwal, Radford, and Klimov}]{schulman2017proximal}
Schulman, J.; Wolski, F.; Dhariwal, P.; Radford, A.; and Klimov, O. 2017.
\newblock Proximal policy optimization algorithms.
\newblock \emph{arXiv preprint arXiv:1707.06347}.

\bibitem[{Silva et~al.(2020)Silva, Gombolay, Killian, Jimenez, and Son}]{silva2020optimization}
Silva, A.; Gombolay, M.; Killian, T.; Jimenez, I.; and Son, S.-H. 2020.
\newblock Optimization methods for interpretable differentiable decision trees applied to reinforcement learning.
\newblock In \emph{International conference on artificial intelligence and statistics}, 1855--1865. PMLR.

\bibitem[{Sutton et~al.(1999)Sutton, McAllester, Singh, and Mansour}]{sutton1999policy}
Sutton, R.~S.; McAllester, D.; Singh, S.; and Mansour, Y. 1999.
\newblock Policy gradient methods for reinforcement learning with function approximation.
\newblock \emph{Advances in neural information processing systems}, 12.

\bibitem[{Topin et~al.(2021)Topin, Milani, Fang, and Veloso}]{topin2021iterative}
Topin, N.; Milani, S.; Fang, F.; and Veloso, M. 2021.
\newblock Iterative bounding mdps: Learning interpretable policies via non-interpretable methods.
\newblock In \emph{Proceedings of the AAAI Conference on Artificial Intelligence}, volume~35, 9923--9931.

\bibitem[{Vos and Verwer(2023)}]{vos2023optimal}
Vos, D.; and Verwer, S. 2023.
\newblock Optimal Decision Tree Policies for Markov Decision Processes.
\newblock In Elkind, E., ed., \emph{Proceedings of the Thirty-Second International Joint Conference on Artificial Intelligence, {IJCAI-23}}, 5457--5465. International Joint Conferences on Artificial Intelligence Organization.
\newblock Main Track.

\end{thebibliography}

\clearpage

\appendix

\section{Environment Descriptions}

We used the Gymnax~\cite{gymnax2022github} implementation of the classic control environments and ported the MDPs from OMDT~\cite{vos2023optimal} to Gymnax. We provide a brief description of each environment below. The environments are described in more detail in their referenced websites and papers:
\begin{itemize}
    \item Acrobot-v1\footnote{\url{https://gymnasium.farama.org/environments/classic\_control/acrobot/}}: Two attached arms that are fixed on one end. By applying torque on the fixed point, the agent should swing the tip of the second arm above a certain height as quickly as possible.
    \item CartPole-v1\footnote{\url{https://gymnasium.farama.org/environments/classic\_control/cart\_pole/}}: A pole balanced on a cart that can move to the left or right. The agent gets 1 reward for every time step that the pole stands upright, and the environment ends when the pole falls or the cart leaves the screen limits.
    \item CartPoleSwingup~\cite{ha2017evolving}: Similar to CartPole-v1, the pole hangs beneath the cart instead of the pole starting upright. This environment is significantly harder as the agent needs to find a strategy to swing the pole up and then control the pole to stay upright.
    \item Pendulum-v1 (discrete)\footnote{\url{https://gymnasium.farama.org/environments/classic\_control/pendulum/}}: An arm that is fixed on one end. By applying torque in the clockwise or counter-clockwise direction, the agent must balance the arm upright. \textbf{We modified this environment by discretizing the continuous action into two actions, `torque left' and `torque right,' that apply maximum torque in the given direction.}
\end{itemize}
For more information about the environments below, please see the OMDT paper~\cite{vos2023optimal}:
\begin{itemize}
    \item Blackjack: A simplified version of the classical card game in which the agent must beat the dealer by deciding whether to draw cards for a higher value and risk or stop drawing.
    \item Frozenlake4x4, 8x8 and 12x12: A maze-like game played on a 2D grid. The agent must get from the start tile to the finish while avoiding holes. Instead of deterministic moves `up', `right', `down', and `left', these actions send the agent in a random, but not opposite, direction. E.g. `left' sends the agent `down', `left', or `up' with equal probability. The maps 4x4, 8x8, and 12x12 increase in size and thus in sparsity of the reward signal for completing the maze.
    \item InventoryManagement: A shop that needs to decide how many items to order with uncertain amounts of customers arriving each day. The agent must balance costs for ordering, missing profits, and storing items.
    \item Navigation3D: The agent needs to move from one end of a 3D grid to the opposite side, with each voxel having its fixed random chance of making the agent disappear. By avoiding bad voxels, the agent can find a strategy to reach the goal state often. \textbf{We modified this environment by reducing the probability of disappearing in each state by a factor of 5 since, in the original environment, no reinforcement learning algorithm managed to learn good policies.}
    \item SystemAdministrator 1, 2 and Tree: A network of computers where every timestep computers have a small probability of crashing, this probability increases when neighboring computers in the network have crashed. The system administrator can reboot one computer at each timestep to fix the crash. Networks 1 and 2 are randomly generated topologies, and `tree' is shaped like a tree.
    \item TictactoeVsRandom: The classic game of tic-tac-toe where players place crosses and circles in a 3x3 grid, and the player that forms a line of 3 consecutive symbols wins the game. In this environment, the opponent is modeled by a player that makes random legal moves.
    \item TigerVsAntelope: The agent controls a tiger that wins the game when it collides with the antelope on a 5x5 grid. The antelope makes random moves away from the tiger.
    \item TrafficIntersection: In an intersection with cars from 2 sides, the operator must decide when to toggle the side with green traffic lights. The operator is rewarded for passing cars through the intersection and penalized for toggling lights and having waiting cars.
    \item Xor: An MDP version of the XOR supervised learning task. The agent is initialized on a random spot on a 2D plane and gets rewarded for taking the action according to an XOR-like function.
\end{itemize}

\section{Hyperparameters}

We have listed the most important hyperparameter settings used in our performance comparison in Tables~\ref{tab:methods-params}~and~\ref{tab:dtpo-params}. When training VIPER, we use the specified DQN parameters for the teacher. More details can be found in our code.

\begin{table}[tb]
    \centering
    \begin{tabular}{lr}
        \toprule
        \textbf{hyperparameter} & \textbf{value} \\
        \midrule
        \multicolumn{2}{c}{DQN} \\
        \midrule
        total timesteps & 10,000,000 \\
        learning rate & 2.5e-4 \\
        experience replay buffer size & 10000 \\
        future discount factor & 0.99 \\
        target network update rate & 1.0 \\
        target network update frequency & 500 \\
        batch size & 128 \\
        starting exploration probability & 1.0 \\
        final exploration probability & 0.05 \\
        fraction of timesteps until final exploration & 0.5 \\
        steps until starting learning & 10000 \\
        train frequency & 10 \\
        activation function & ReLU \\ 
        layer sizes & 120, 84 \\ 
        Adam optimizer $\beta_1$ & 0.9 \\
        Adam optimizer $\beta_2$ & 0.999 \\
        Adam optimizer $\epsilon$ & 1e-8 \\
        \midrule
        \multicolumn{2}{c}{VIPER} \\
        \midrule
        future discount factor & 0.99 \\
        number of rollouts per batch & 10 \\
        maximum samples when training tree & 200,000 \\
        maximum number of trees & 80 \\
        fraction of train samples & 0.8 \\
        test rollouts per tree & 50 \\
        maximum number of leaf nodes & 16 \\
        \midrule
        \multicolumn{2}{c}{PPO} \\
        \midrule
        total timesteps & 10,000,000 \\
        learning rate & 2.5e-4 \\
        timesteps per iteration & 128 \\
        parallel environments & 4 \\
        batches per epoch & 4 \\
        epochs per iteration & 4 \\
        future discount factor & 0.99 \\
        $\lambda$-returns trace decay & 0.95 \\
        PPO clipping value ($\epsilon$) & 0.2 \\
        entropy coefficient & 0.01 \\
        value function coefficient & 0.5 \\
        maximum gradient clipping norm & 0.5 \\
        activation function & tanh \\ 
        actor layer sizes & 64, 64 \\ 
        critic layer sizes & 64, 64 \\ 
        Adam optimizer $\beta_1$ & 0.9 \\
        Adam optimizer $\beta_2$ & 0.999 \\
        Adam optimizer $\epsilon$ & 1e-5 \\
        \bottomrule
    \end{tabular}
    \caption{Overview of the hyperparameter settings used for DQN, VIPER and PPO. VIPER uses the DQN as a teacher.}
    \label{tab:methods-params}
\end{table}

\begin{table}[tb]
    \centering
    \begin{tabular}{lcr}
        \toprule
        \textbf{hyperparameter} & \textbf{symbol} & \textbf{value} \\ 
        \midrule
        \multicolumn{3}{c}{DTPO decision tree settings} \\
        \midrule
        timesteps per iteration & T & 10,000 \\
        learning rate & $\eta$ & 1 \\
        future discount factor & $\gamma$ & 0.99 \\
        $\lambda$-returns trace decay & $\lambda$ & 0.95 \\
        maximum iterations & $N$ & 1,500 \\ 
        maximum number of leaf nodes & - & 16 \\
        \midrule
        \multicolumn{3}{c}{DTPO value function neural network} \\
        \midrule
        batch size & B & 64 \\
        number of epochs & E & 4 \\
        learning rate & - & 2.5e-4 \\
        activation function & - & tanh \\ 
        layer sizes & - & 64, 64 \\ 
        Adam optimizer $\beta_1$ & - & 0.9 \\
        Adam optimizer $\beta_2$ & - & 0.999 \\
        Adam optimizer $\epsilon$ & - & 1e-8 \\
        \bottomrule
    \end{tabular}
    \caption{Overview of the hyperparameters used for DTPO in the results section and their symbols as they appeared in this paper.}
    \label{tab:dtpo-params}

    \vspace{1.7cm}

    \setlength{\tabcolsep}{5.5pt}
    \begin{tabular}{@{}l|rrrr@{}}
    \toprule
    \textbf{environment} & \textbf{VIPER} & \textbf{DTPO} & \textbf{DQN} & \textbf{PPO} \\
    \midrule
Acrobot-v1 & 7.0 & 0.6 & 6.8 & 1.3 \\
Blackjack & 7.8 & 1.4 & 6.8 & 1.2 \\
CartPole-v1 & 7.1 & 0.6 & 6.7 & 1.3 \\
CartPoleSwingup & 8.0 & 0.6 & 6.8 & 1.2 \\
Frozenlake12x12 & 6.7 & 0.6 & 6.6 & 1.2 \\
Frozenlake4x4 & 6.5 & 0.6 & 6.5 & 1.2 \\
Frozenlake8x8 & 6.7 & 0.6 & 6.5 & 1.2 \\
InventoryManagement & 3.9 & 1.5 & 3.3 & 1.4 \\
Navigation3D & 7.8 & 0.6 & 6.6 & 1.2 \\
PendulumBangBang & 7.0 & 0.6 & 6.7 & 1.2 \\
Sysadmin. 1 & 7.7 & 1.1 & 6.6 & 1.2 \\
Sysadmin. 2 & 7.6 & 1.2 & 6.6 & 1.2 \\
Sysadmin. Tree & 7.6 & 0.8 & 6.6 & 1.3 \\
TictactoeVsRandom & 7.5 & 2.2 & 7.5 & 1.6 \\
TigerVsAntelope & 7.5 & 1.6 & 6.7 & 1.5 \\
TrafficIntersection & 7.7 & 1.1 & 6.7 & 1.5 \\
Xor & 7.9 & 0.6 & 6.6 & 1.4 \\
    \bottomrule
    \end{tabular}
    \caption{Median runtimes in hours over 3 runs. The DQN training time is included in VIPER's runtime.}
    \label{tab:runtime-comparison}
\end{table}

\section{Runtimes}
The median runtimes in hours of our experimental comparison are given in Table~\ref{tab:runtime-comparison}. DTPO and PPO usually run in at most 2 hours, while training the DQN and VIPER can take 8 hours. However, we did not optimize our implementations and expect VIPER and DQN to benefit significantly if they were further integrated with the optimized Gymnax~\cite{gymnax2022github} environments. Most of the runtime of VIPER is spent training the DQN model.

\section{Regression Tree Learning}

For regression tree learning, we rely on the standard implementation of Scikit-learn\footnote{\url{https://scikit-learn.org/stable/modules/generated/sklearn.tree.DecisionTreeRegressor.html}}. We also provide pseudocode in Algorithm~\ref{alg:regression-tree} for completeness.

\begin{algorithm*}[tb]
    \caption{Greedy Multi-Output Regression Tree Learning}\label{alg:regression-tree}
    \begin{algorithmic}[1]
        \State \textbf{Input:} the matrix of observations $X \in \mathbb{R}^{n \times m}$ and the targets $Y \in \mathbb{R}^{n \times k}$ where $k$ is the number of outputs
        \Function{FitRegressionTree}{X, Y}
            \If{stopping criterion (e.g. maximum number of nodes, pure leaf)}
                \State \Return \textsc{Leaf}$\left(\frac{1}{|Y|} \sum_i Y_{i1}, \frac{1}{|Y|} \sum_i Y_{i2}, ..., \frac{1}{|Y|} \sum_i Y_{ik}\right)$
            \EndIf
            \State define $G$ as the weighted mean of the outputs' mean squared errors, where $\overline{Y}_{*c}$ is the mean of $Y$'s output $c$:
            \State $G(Y^l, Y^r) = \frac{|Y^l|}{|Y^l| + |Y^r|} \frac{1}{k} \sum_{c = 1}^k \frac{1}{|Y^l|} \sum_{i = 1}^{|Y^l|} (Y^l_{ic} - \overline{Y^l}_{*c})^2 + \frac{|Y^r|}{|Y^l| + |Y^r|} \frac{1}{k} \sum_{c = 1}^k \frac{1}{|Y^r|} \sum_{i = 1}^{|Y^r|} (Y^r_{ic} - \overline{Y^r}_{*c})^2$
            \State $j^*, v^* = \text{argmin}_{j, v} G(\{Y_i : X_{ij} \leq v\}, \{Y_i : X_{ij} > v\})$
            \State $T_l \gets $ \Call{FitRegressionTree}{$X_i, Y_i$ $ : X_{ij^*} \leq v^*$}
            \State $T_r \gets $ \Call{FitRegressionTree}{$X_i, Y_i$ $ : X_{ij^*} > v^*$}
            \State \Return \textsc{Node}$(j^*, v^*, T_l, T_r)$
        \EndFunction
    \end{algorithmic}
\end{algorithm*}

\end{document}